\documentclass[journal]{IEEEtran} 
\usepackage{cite}

\ifCLASSINFOpdf
   \usepackage[pdftex]{graphicx}
\else
\fi
\usepackage{multicol, blindtext}

\usepackage{amsmath,amssymb,amsfonts}
\usepackage[margin=2cm]{geometry}
\usepackage{adjustbox}
\usepackage{multirow}
\usepackage{ctable} 
\usepackage{array}

\usepackage{flushend,epstopdf}

\begin{document}
%
\title{Coordinates-based Resource Allocation Through\\ Supervised Machine Learning}
%
%
%

\author{Sahar~Imtiaz,~\IEEEmembership{Student~Member,~IEEE,}
        Sebastian~Schiessl,~\IEEEmembership{Member,~IEEE.} 
        Georgios~P.~Koudouridis,~\IEEEmembership{Member,~IEEE.}
        and~James~Gross,~\IEEEmembership{Senior Member,~IEEE.}%
\thanks{S. Imtiaz and J. Gross are with the Division
of Information Science and Engineering, KTH Royal Institute of Technology, Stockholm, Sweden, 10044. (e-mail: sahari@kth.se, james.gross@ee.kth.se)}%
\thanks{S. Schiessl is with u-blox Athens S.A., 15125 Maroussi, Greece. (e-mail: sebastian.schiessl@u-blox.com)}%
\thanks{G. P. Koudouridis is with Wireless Systems Lab, Stockholm Research Center, Huawei Technologies Sweden AB, Kista, Stockholm, Sweden. (email: george.koudouridis@huawei.com)}%
\thanks{Manuscript received - May 1, 2020}}

%
%

\markboth{ }
{Shell \MakeLowercase{\textit{et al.}}: Bare Demo of IEEEtran.cls for IEEE Journals}
%



\maketitle
\begin{abstract}
Appropriate allocation of system resources is essential for meeting the increased user-traffic demands in the next generation wireless technologies. 
Traditionally, the system relies on channel state information (CSI) of the users for optimizing the resource allocation, which becomes costly for fast-varying channel conditions. 
Considering that future wireless technologies will be based on dense network deployment, where the mobile terminals are in line-of-sight of the transmitters, the terminals' position information provides an alternative to estimate the channel condition. 
In this work, we propose a coordinates-based resource allocation scheme using supervised machine learning techniques, and investigate how efficiently this scheme performs in comparison to the traditional approach under various propagation conditions. 
We consider a simplistic system set up as a first step, where a single transmitter serves a single mobile user. 
The performance results show that the coordinates-based resource allocation scheme achieves a performance very close to the CSI-based scheme, even when the available user's coordinates are erroneous. 
The proposed scheme performs consistently well with realistic-system simulation, requiring only 4~s of training time, and the appropriate resource allocation is predicted in less than 90 $\mu$s with a learnt model of size $<$1 kB. 
\end{abstract}


%
\IEEEpeerreviewmaketitle

\section{Introduction}
\label{sec:1:intro}

Efficient allocation of system resources among the users of a network is the key to improved system performance. 
The resource allocation refers to, for example, the allocation of resource blocks in time or frequency domain, the allocation of transmit power, and/or modulation and coding scheme for payload transmission, etc. to a specific user in the system. The task of determining the appropriate resource allocation relies on the information about the propagation environment acquired by the transmitter(s), as well as on the available computational power in the system. 
Traditionally, the users' channel state information (CSI) available at the transmitters is utilized for resource allocation, where heuristic approaches can be used to optimize the practical implementation~\cite{alternate_exhaustive_search}. 
The CSI acquisition contributes to a performance overhead, which varies depending on the system's user density. 
The last few decades have seen a surge in the user traffic demands, along with immensely increased user density, in a wireless communication system. 
This implies that instantaneous CSI acquisition will incur significant overhead for such systems with high user density where the channel varies frequently due to the propagation environment and user mobility. 
Furthermore, high user density leads to increased computational complexity for efficient resource allocation. 
This arises a need for alternate approaches for resource allocation, such that the overhead for acquiring the propagation environment's information is minimal and at the same time, the required computational complexity to optimize the resource allocation is within certain constraints.
According to a recent survey~\cite{Jiang_30MLSurvey_2020}, several research works in the last decade have focused on the idea of applying various machine learning frameworks for resource allocation in wireless systems. 
In addition to the related work mentioned in~\cite{Jiang_30MLSurvey_2020}, the work in~\cite{TimOShea_DySpan_2017} applies deep neural network for cognitive radio modulation recognition. 
The authors in~\cite{Zanella_MEDHOCNET_2014} combine unsupervised feature learning with supervised classification techniques to design an efficient quality-of-experience-based video admission control and resource allocation scheme. 
Deep reinforcement learning is used in~\cite{Gursoy_RAinCRAN_ICC2017} to propose a power-efficient resource allocation scheme for wireless system based on cloud radio access network architecture. 
All these works highlight the advantages of exploiting machine learning for resource allocation in wireless systems, without compromising the computational capacity of the system.

Besides relying on CSI for optimizing the performance of wireless systems, various research works propose the use of position information in this context. 
Majority of these works discuss position-based schemes for improving the system performance at different layers of the protocol stack \cite[\& references therein]{Survey_Taranto_2014}, \cite{Schmandt_Springer_2000}, but only a handful of those consider position information for resource allocation. 
It should be noted that the position information can be acquired through narrow-band uplink pilots, as opposed to full-band pilots used for CSI acquisition, resulting in a considerably lower overhead. 
\cite{Slock_ISCCSP_2012} provides an overview of the position-based resource allocation to improve the performance of wireless systems, stressing the fact that the true potential of position-based techniques needs to be evaluated by comparison with the CSI-based methods. 
In~\cite{Petteri_Beamforming_2016}, position-related information (i.e. angle-of-arrival) of mobile terminals is utilized for spatial filtering in an ultra-dense network to maximize throughput. 
A similar approach is used in~\cite{Letaief_IEEE_Access_2017}, where the authors propose location-aided beamforming by exploiting the distance between the base station and the mobile relay for beam selection to serve high-speed users.
Other works like~\cite{Cha:2017:CSS:3022227.3022279} and~\cite{Raulefs_Asilomar_2009} consider location-based resource allocation in conjunction with CSI. 
To the best of our knowledge, none of the previous works have considered resource allocation solely based on position coordinates of the mobile terminal. 
Furthermore, most of the above studies do not account for stochastic variations affecting the position-related information, rather assume that the position is perfectly known without considering estimation errors. 
Our recent work~\cite{Sahar_GLOBECOM_2019} presents some initial investigation results about the feasibility of a coordinates-based resource allocation scheme, but the main question about the associated implementation constraints in real-time system is what we address in this work.  

We focus on the simplistic system comprising a single transmitter serving a single mobile terminal, with dominant line-of-sight (LoS) communication link. 
This implies that the channel characteristics of the propagation environment of the mobile terminal is related to its position information. 
We apply supervised machine learning framework to learn this relationship and design a coordinates-based resource allocation scheme that maximizes the transport capacity of the system. 
This study extends our previous work~\cite{Sahar_GLOBECOM_2019} to comprehensively address the applicability of different machine learning frameworks to model and perform coordinates-based resource allocation. 
More specifically, the main contributions of our work are as follows:
\begin{itemize}
    \item We present a detailed description of the coordinates-based resource allocation scheme through machine learning proposed in~\cite{Sahar_GLOBECOM_2019}, and discuss the different possibilities for dataset formulation and the associated challenges. 
    \item We investigate the applicability of the proposed coordinates-based resource allocation scheme using the different dataset formulations. Based on the best possible choice for dataset formulation, we investigate the performance results of the proposed scheme with respect to the stochastic variations in system characterization.  
    \item In terms of implementation constraints, we present an analysis of the time required to train the proposed coordinates-based resource allocation scheme through machine learning. Particularly, we consider realistic system simulation with correlated channels, and determine the training time necessary for the proposed scheme to achieve a system performance comparable to the CSI-based resource allocation scheme. 
\end{itemize}

The rest of the paper is structured as follows: Section~\ref{sec:2:sys_model} describes the system model considered in this work, while Section~\ref{sec:3:ML_&_scheme} discusses the design of the proposed resource allocation scheme, along with the details of the machine learning frameworks used in this work. 
The different dataset formulations together with their analysis are discussed in Section~\ref{sec:4:datasets}. 
Section~\ref{sec:5:results_main_section} presents the results and relevant discussions, along with the analysis of the training time required for real-time implementation of the proposed resource allocation scheme. 
Section~\ref{sec:6:conclusion} concludes the paper. 

\section{System Model}
\label{sec:2:sys_model}

In this section we first describe the basic system model, followed by a formal definition of the resource allocation problem. 
This problem definition is based on an optimization function which aims at maximizing the transport capacity of the system.

\subsection{Basic System Model}
\label{subsec2:1:basic_sys_model}
We focus on the downlink communication between a single base station (BS), and a single user terminal. 
We assume that the system operates on time frames of duration $T_{\textrm{f}}$ [ms]. 
OFDM waveform is considered for payload transmission, with the BS utilizing a bandwidth $W$ spanning a number of sub-carriers $N$. 
The BS is equipped with $A_{\textrm{r}}$ transmit antennas,  collectively operated with a constant transmit power denoted by $P_{\textrm{r}}$ [W]. 
The user terminal is assumed to have $A_{\textrm{u}}$ receive antennas.

We consider a time-varying wireless communication channel between the BS and the terminal, subject to pathloss, shadowing and fading. 
Let us denote by $\pmb{H}(t,n)$ the MIMO channel matrix between all transmit and receive antenna elements at time $t$ and for sub-carrier $n$. 
We assume that the MIMO channel stays constant during a single time frame and that the user terminal is not exposed to any interference. 
With these assumptions in mind, if the BS applies a transmit beam $\pmb{v}$ at time $t$ to transmit the symbol $s(t,n)$ over sub-carrier $n$, the received signal at the terminal, when the terminal applies a receive filter $\pmb{u}$, will be given by:
\begin{align}
\label{eq:1:rx_signal}
    y(t,n) = \sqrt{P_{\textrm{r}}} \cdot (\pmb{u}(t))^{\dagger} \cdot \pmb{H}(t,n) \cdot \pmb{v}(t) \cdot s(t,n) + z,
\end{align}
\noindent
where $z$ represents the additive white Gaussian noise, and $(\pmb{u}(t))^{\dagger}$ represents the Hermitian of $\pmb{u}(t)$. 
Based on the received signal in \eqref{eq:1:rx_signal}, the signal-to-noise ratio (SNR) for the terminal served by BS at time $t$ for sub-carrier $n$, with noise power $\sigma_z^2$, is given by:
\begin{align}
    \label{eq:2:SNR_eq}
    \gamma(t,n) = \frac{P_{\textrm{r}} \cdot |(\pmb{u}(t))^{\dagger} \cdot \pmb{H}(t,n) \cdot \pmb{v}(t) |^2} {\sigma_z^2}.
\end{align}

For the transmission of payload to the receiver, the BS applies a modulation and coding scheme (MCS) $m$ over all the $N$ sub-carriers. 
The chosen MCS is applied uniformly for all the $N$ sub-carriers, meaning that the system does not feature adaptive MCS per sub-carrier. 
We assume a full-buffer state at the BS with respect to the terminal. 
Therefore, the payload transmission based on the choice of MCS carries a maximum possible number of bits, denoted by $b(m(t))$. 
Due to noise, the transmitted bits can be received erroneously at the terminal, which can be modelled by an error function $\epsilon$. 
This results in transport capacity (or goodput) $\mathcal{T}(t)$ at the terminal, by which we measure the performance, and is given by:
\begin{align}
    \label{eq:3:goodput_eq}
    \mathcal{T}(t) = (1-\epsilon) \cdot b(m(t)).
\end{align}

\subsection{Resource Allocation and the Position Information}
\label{subsec2:2:RA_&_position}

We consider the maximization of transport capacity as the optimization problem in this work. 
The transport capacity at the terminal depends on the allocation of available system resources between the BS-terminal pair. 
We denote the resource allocation by $\mathfrak{r}(t)$, which comprises the following: (a) the transmit beam $\pmb{v}(t)$ applied by the BS, (b) the receive filter $\pmb{u}(t)$ applied by the terminal, and (c) the MCS $m(t)$ chosen by the BS. 
The transmit beam $\pmb{v}(t)$ and the receive filter $\pmb{u}(t)$ is chosen from the finite sets $\mathbb{V}$ and $\mathbb{U}$, respectively. 
The sets $\mathbb{V}$ and $\mathbb{U}$ are pre-determined using geometric beamforming, with an angular separation $\theta^{\circ}$, at both the BS and the user terminal, respectively. 
The MCS value $m(t)$ is applied from the finite set $\mathbb{M}$, that is based on the link-to-system mapping given in \cite{afifi2013radio}. 
Since the downlink communication between a single BS and user terminal is the focus of our work, we allocate all the time and frequency resources, as well as the full transmit power, to the single terminal. 
Based on the above description, the optimization problem for resource allocation per downlink frame can be stated as: 
\begin{align}
    \label{eq:4:RA_prob}
    \max_{\mathfrak{r}(t)} \mathcal{T}(t) =  (1 - \epsilon) \cdot b(m(t)). 
\end{align}

Traditionally, the resource allocation problem is solved based on the CSI of the BS-terminal pair, which requires the CSI to be perfectly known at the BS. 
This means that the system has to apply full-bandwidth signals, with frequent signaling, but in reality the signaling bandwidth is limited and signaling resources are scarce. 
Due to this limitation and scarcity, CSI estimation would result in a lower transmission rate than required per user, in addition to an outdated CSI information. 
An outdated CSI results in a highly inefficient resource allocation and, consequently, a deterioration of the system performance. 
This is prevalent in scenarios with many users, and fast channel changes due to user mobility and high density of scatterers. 
To mitigate these problems, an alternate approach is needed for efficient resource allocation that maximizes the system performance.  

In this work, we consider systems where an estimate of the terminal's position can be obtained at the BS in addition to CSI acquisition. 
This position estimate can be determined, for example, by Kalman filtering of the direction-of-arrival and time-of-arrival of the specifically sent positioning beacons in the uplink \cite{Petteri_Beamforming_2016}. 
These positioning beacons are in fact narrow-band signals, which pose significantly lesser overhead compared to CSI estimation beacons for high user density scenarios, as mentioned in \cite{EURASIP_Sahar_2018}.
Let $\pmb{p}(t)$ denote the true position coordinates of the terminal at time $t$, while $\hat{\pmb{p}}(t)$ denote the estimate of position coordinates of the terminal. 
Assuming that dominant LoS link exists between the BS and the terminal, with no interference exposed at the terminal, the channel characteristics remain fairly constant, and are related to the terminal's position estimate known at the BS. 
However, this relationship is affected by the fact that the position of terminal can not be accurately known at the BS at all times, as well as by the presence of scatterers in the propagation area. 
In this work, we first investigate under which conditions a relationship between the position estimate of terminal and the channel state exists, and if so, how can this relationship be exploited to maximize the transport capacity $\mathcal{T}(t)$? 
This question forms the basis of our research problem, which is presented in the next sub-section. 

\subsection{Problem Statement}
\label{subsec2:3:prob_statement}

As mentioned before, maximum transport capacity $\mathcal{T}$ is achieved when resource allocation is determined optimally based on the perfectly known CSI at the BS. 
However, depending on the propagation scenario, the instantaneous CSI acquisition can be costly or the available CSI estimates can be outdated, both of which are detrimental to CSI-based resource allocation. 
For the propagation scenarios where dominant LoS exists between the BS-terminal pair, the relationship between the estimated position coordinates of the terminal $\hat{\pmb{p}}(t)$ and the downlink channel state can be exploited to determine the resource allocation for BS-terminal pair for solving the optimization problem \eqref{eq:4:RA_prob}. 
Intuitively, with the estimated position of the mobile terminal $\hat{\pmb{p}}(t)$ available at the BS, geometric beamforming can be applied to determine the resource allocation $\mathfrak{r}(t)$. 
This means that the transmit beam $\pmb{v}(t)$ and the receive filter $\pmb{u}(t)$ are determined based on $\hat{\pmb{p}}(t)$, whereas the MCS $m(t)$ can be determined based on the distance between the BS and the terminal. 
But this approach suffers from the inaccurate position information availability at the BS from time to time, in addition to being affected by the presence of scatterers in the propagation area. 
Furthermore, the geometric approach suffers from the antenna radiation profiles and the antenna orientation, at both the BS and the terminal. 
In this work, we apply supervised machine learning to design a coordinates-based resource allocation scheme to solve the capacity maximization problem. 
In particular, we discuss how supervised machine learning can be used to determine the resource allocation $\mathfrak{r}(t)$ from terminal's position estimates $\hat{\pmb{p}}(t)$, and how will such a coordinates-based resource allocation scheme be implemented in a wireless communication system.  
Furthermore, we will investigate the computational cost associated with the implementation of the proposed scheme in a realistic system setup. 

\section{Coordinates-based Resource Allocation Using Machine Learning}
\label{sec:3:ML_&_scheme}

In this section, we first present the design and working of the coordinates-based resource allocation scheme. We also discuss some challenges related to the design of the proposed scheme, and the different possible solutions we considered. Afterwards, we outline the different supervised machine learning algorithms used for coordinates-based resource allocation, and the motivation for their choice. 

\subsection{Design and Working of the Proposed Scheme}
\label{subsec3:1:proposed_scheme}

\begin{figure}[!h]
\centering
\includegraphics[width=8cm]{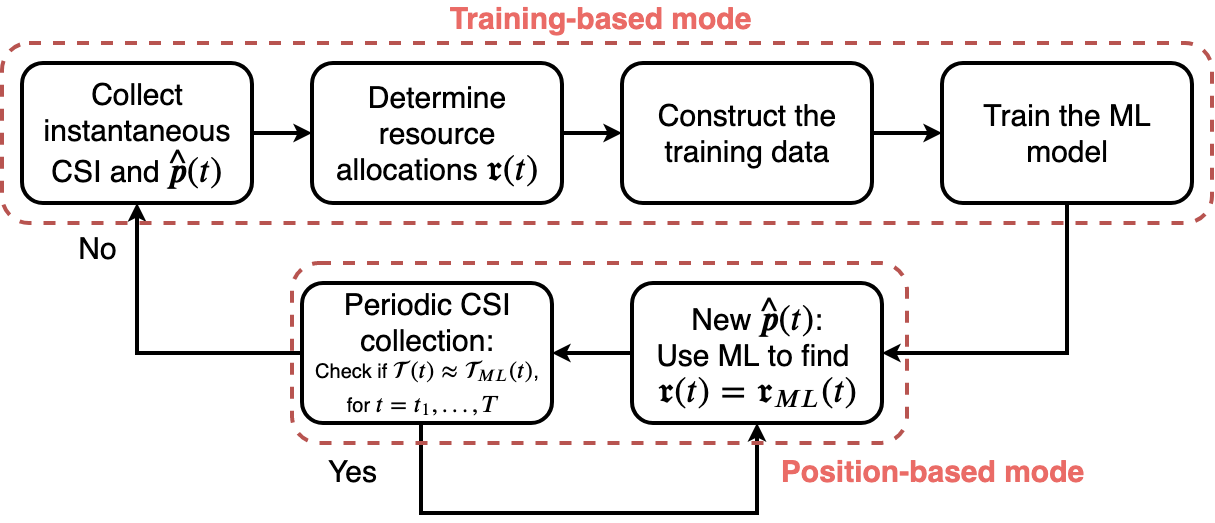}
\caption{Working of the Coordinates-based Resource Allocation Using Machine Learning (ML)}
\label{fig:1:scheme}
\end{figure}

As mentioned before, we apply supervised machine learning framework for designing the coordinates-based resource allocation scheme. 
This implies that the samples collected for training the machine learning framework will comprise of input parameters and an output, which will be predicted by the learnt model. 
Fig.~\ref{fig:1:scheme} shows the design and working of the coordinates-based resource allocation scheme. The scheme consists of two modes, namely the training-based mode and the position-based mode. 
In the beginning, the system operates in the training-based mode, where the data collection process happens simultaneously with the CSI-based resource allocation. 
In this mode, both the estimate of terminal's coordinates $\hat{\pmb{p}}(t)$ as well as its CSI are collected by the system for a period of time to construct the training samples. 
This information is processed offline, where the CSI for each collected sample is used to determine the resource allocations $\pmb{\mathfrak{r}}(t)$ that maximize the system's transport capacity. 
Once all the samples are processed, the training dataset is formulated by associating each position estimate $\hat{\pmb{p}}(t)$, the input, with the corresponding resource allocations $\pmb{\mathfrak{r}}(t)$, the output. 
These samples are used to train the supervised machine learning frameworks, and the corresponding learning models are then used for prediction by the system. 

After the training process is complete, the system operates in position-based mode. 
In this mode, only the terminal's position coordinate for each time frame is assumed to be known. 
This estimate is then passed to the trained machine learning model for determining the resource allocation based on the model's prediction. 
To ensure efficient performance, the predictions from the learnt model need to be checked with the baseline CSI-based resource allocation from time to time. 
In case the goodput computed by the predicted resource allocations is not inline with the one given by the CSI-based resource allocation, the system switches back to the training-based mode to retrain the machine learning model. 
The exact modelling of this mode-switching process is out of the scope of this work.
However, we provide an intuition about the time needed to collect a sufficient amount of training samples, and to train the machine learning model, for a stable learning performance towards the end of this paper.

With respect to the machine learning frameworks, the major design challenge relates to the representation of inputs and outputs in the dataset. 
In this case, two basic representations for input variables are possible: One is to treat the propagation scenario in the form of a binary-coded image, where the position estimate of the terminal is marked with a 1, while the rest of the image is coded as 0's. 
The other approach is to use the coordinates of the estimated terminal's position as the input vector for learning framework. 
The image-based representation is specially suited for neural network-based learning methods, but due to the fact that the coded vector will be highly sparse, as it will indicate the position estimate of only a single terminal, a huge amount of data samples will be required to train the neural network appropriately. 
The collection of huge amounts of data samples yields this data representation impractical for implementation. 
In contrast to the image-based representation, the other approach uses the estimated coordinates of the terminal itself as input, which are stored as floating point numbers in the system. 
However, this representation entails a low-dimensional input vector, and therefore, deep learning architectures can not be used with this data representation. 
We choose the low-dimensional input vector representation for other learning frameworks, due to feasibility of this data representation in a real-time setup. 
Inspired by this input representation, we choose to encode $\mathfrak{r}(t)$ as a binary string, where the different parts of the string encode the individual resource variables' information. 

Besides data representation, another issue relates to the dataset formulation itself, considering the association between input and output variables, or classes. 
In our case, such a data formulation is unique in nature since a single position estimate can be associated with multiple resource allocations, or classes, that maximize the transport capacity. 
In general, the datasets used for machine learning have a unique relationship between inputs and outputs, but this is not the case for our set up. 
With these restrictions in mind, different methods for dataset formulation can be considered. 
One formulation is based on designing the dataset for binary classification problem, but this implies that the output variable can take one of the two possible values. 
This is not possible for the choice of resource allocation as an output variable, and hence, rules out the usage of support vector machines algorithm, which is fundamentally a binary classifier \cite{bishop2006pattern}. 
In this work, we consider the dataset formulation based on single output variable per sample, where the details of the different possible formulations can be found in Section~\ref{subsec4:3:data_formulation}.
With the above choice of data representation and dataset formulation, we now present in detail the machine learning frameworks used in our work. 

\subsection{The Machine Learning Frameworks}
\label{subsec3:3:ML}

The supervised machine learning domain mainly comprises the following well-known algorithms, ranging from lowest to highest possible complexity: K-nearest neighbor (KNN), support vector machines, random forest (RF) and neural network.
Based on the representation choice of the input and output variables in the dataset formulation, we use K-nearest neighbors and Random Forests algorithm as the machine learning frameworks in our work. 
The details of KNN and RF algorithms are mentioned below.

\subsubsection{K-Nearest Neighbor Algorithm}
\label{subsubsec3:1:KNN}
KNN is the simplest machine learning framework \cite{duda1995pattern} and does not build an explicit model for predicting the classes. 
Instead, the whole training dataset itself is used for class prediction.  
For a given sample of the test dataset, the KNN first determines the K samples in the training data that are closest to the test data sample. 
Then it performs a majority vote on the class associated with the K nearest neighbors to predict the outcome for the given test data sample. 
In the context of the coordinates-based resource allocation, given a training dataset with sufficient sampling of the terminal's coordinates, the KNN can capture the spatial relationship between the terminal's position coordinates and the respective resource allocation quite accurately.

\subsubsection{Random Forest Algorithm}
\label{subsubsec3:2:RF}
Random Forest \cite{Breiman_RF} is a complex supervised learning algorithm, which builds a learning model for class prediction, as opposed to KNN. 
The model consists of an ensemble of randomized binary decision trees, where each decision tree is constructed using a dataset consisting of samples taken from sampling with replacement on the available training data. 
An individual sample in the training data is called an input feature vector, and consists of the input features $\pmb{\mathfrak{f}}$ along with the output variable(s). 
Each decision tree in the forest consists of a root node, several interior nodes and terminal leaf nodes. 
The thresholds for each node are determined based on a subset of randomly selected input features $\pmb{\mathfrak{f}}'$, which induces randomness in each decision tree of the RF model. 
Overall, $\Omega_{\textrm{t}}$ number of trees are constructed, where each tree is either grown to a maximum depth of $\Omega_{\textrm{d}}$ or till all the classes are perfectly separated. 
The predicted class is based on a majority vote on the classes predicted by all the trees in the forest. 

For the coordinates-based resource allocation, whenever a new estimate of terminal's position is available to the RF model, it is parsed through all the trees in the forest. 
The resource allocation prediction is made by taking the mode of the resource allocations, i.e. the classes, predicted by all the trees in the forest. 
The ensemble of decision trees provides robustness to the learnt model, and therefore, RF is robust to the noisy inputs compared to other machine learning frameworks. 
This property makes the choice of RF even more attractive for the cases where noisy estimates of terminal's position coordinates are available for determining the resource allocation that maximize the system performance. 
The randomness introduced by random selection of input features for constructing an individual tree prevents the RF algorithm from over-fitting on the training dataset. 
Due to these reasons, RF algorithm is expected to perform better than KNN, typically for the cases when erroneous estimates of terminal's coordinates are available or when the propagation scenario involves randomness in the channel between BS-terminal pair. 

This section discussed the design of coordinates-based resource allocation scheme, along with the challenges involved in designing the proposed scheme tailored for supervised machine learning frameworks.
Since resource allocation problem is a multi-class classification problem, therefore, support vector machines is not considered in this work for machine learning. 
Furthermore, based on the input data representation, we rule out the usage of neural network for designing the proposed scheme. 
In the next section, we discuss the specific models used to generate the training datasets, followed by the details of different dataset formulations used in our work.

\section{The Datasets}
\label{sec:4:datasets}
In this section, we will first mention the different models considered for simulating the propagation scenario, specific to the system model described in Section~\ref{sec:2:sys_model}. Next we present the methodology for simulating the channel model to generate the datasets. Afterwards, we discuss the different dataset formulations to solve the coordinates-based resource allocation problem using machine learning. At the end, we will present the analysis of datasets for some baseline propagation scenarios. 

\subsection{Scenario Description}
\label{subsec4:1:model_sel}

\begin{figure}[h]
\centering
\includegraphics[width=7cm]{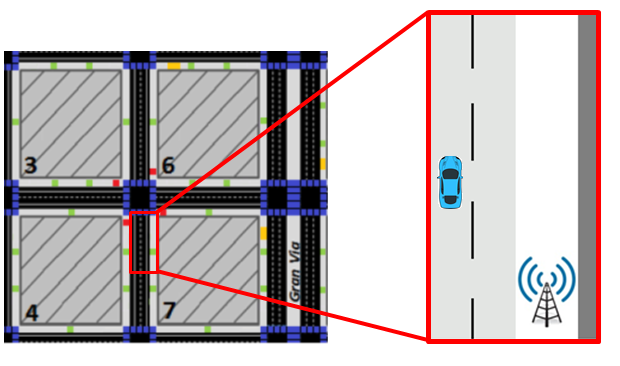}
\caption{The simulation scenario for generating different datasets.}
\label{fig:2:sim_scenario}
\end{figure}

We consider a small street section of $6\times25$ m$^2$ with a single BS serving a single mobile terminal as the propagation scenario, shown in Fig.~\ref{fig:2:sim_scenario}. 
The BS is placed at the right lower end of the street, 3~m off-roadside. 
The mobile terminal is placed randomly over the street (random-drop), with uniform distribution over the entire street section. 
We define a parameter $\rho$ [/m$^2$] to specify the maximum density of scatterers in the considered propagation scenario. 
As an example, if $\rho~\leq$ 0.05/m$^2$, up to 5 scatterers will be randomly placed in the propagation environment. 
The placement as well as the number of scatterers will vary for each random-drop of the terminal. 
The BS is equipped with $A_{\textrm{r}} = \{8, 4\}$ transmit antennas, while the terminal has $A_{\textrm{u}} = \{2, 1\}$ receive antennas.
Each antenna element at both the BS and the terminal is a Hertzian dipole, and forms a uniform linear array oriented along the $x-$axis. 
The BS antenna elements are collectively operated with a power of $P_{\textrm{r}} = $ 1$\mu$W, and are placed at a height of 10 m from the ground. 
The terminal antennas are at a height of 1.5 m from the ground, which remains constant for all the simulation scenarios. 
The system operates on a center frequency $f_{\textrm{c}}$ of 3.5 GHz over a bandwidth $W = $ 200 MHz.
The transmission time interval of the system is set to $T_{\textrm{f}} = $ 0.2 ms. 

To calculate the transport capacity $\mathcal{T}(t)$ in \eqref{eq:4:RA_prob}, we apply the link-to-system mapping error function for $\epsilon$,  $e(\mathfrak{m}(t),\gamma_{\textrm{eff}}(t))$, that emulates the erroneous reception of the transmitted bits at the receiver. 
Here, $\gamma_{\textrm{eff}}$ is the exponential effective SNR mapping \cite{tuomaala2005effective}, which converts the SNR value $\gamma(t,n)$ per sub-carrier into an equivalent SNR over all the $N$ sub-carriers, with respect to the considered communication scenario. 
The link-to-system mapping error function provides an error rate that is specific to a range of $\gamma_{\textrm{eff}}$ values for a given MCS $m(t)$. 
Hence, the link-to-system model $e(m(t),\gamma_{\textrm{eff}}(t))$ represents the relationship of the block error rate, effective SNR, MCS as well as the transmitted payload size.

In terms of the system resources, the sets of transmit beams $\mathbb{V}$ and receive filters $\mathbb{U}$ are determined using geometric beamforming, with an angular separation of $\theta =$ 3$^{\circ}$ and 12$^{\circ}$, respectively. 
The finite set of MCS values $\mathbb{M}$ comprises 15 different values, and is based on the link-to-system mapping in \cite{afifi2013radio}. 
The optimal resource allocations (RAs) $\pmb{\mathfrak{r}}(t)$ are determined by solving Problem~\eqref{eq:4:RA_prob} through exhaustive search. 

Another important parameter is the error in the position estimate for the system model presented in section~\ref{subsec2:1:basic_sys_model}. 
The position estimation error $(\pmb{p}(t)-\hat{\pmb{p}}(t))$ is modelled as a Gaussian zero-mean random variable with variance $\sigma^2$. 
With respect to the communication system, this position error depends on the accurate estimation of the direction of arrival and the time of arrival parameters, where the former depends on the antennas' geometry, while the latter is related to the pathloss between the BS-terminal pair. 
In addition to the aforementioned modelling parameters, the utmost important is choice of the channel model, which we present in the next sub-section. 

\subsection{The Channel Model}
\label{subsec4:2:channel_model}

We resort to simulations for generating the datasets to determine the resource allocation using estimated position coordinates of the terminal to solve the optimization problem \eqref{eq:4:RA_prob}. 
To the best of our knowledge, only the received signal strength related information is available in the publicly accessible traces on various open-source platforms, with no details about the position information of the terminal. 
Therefore, one of the major challenges for data generation is to choose the simulation models that emulate the real-time measurements as closely as possible. 
Hence, we utilize the ray-tracer channel model~\cite{METIS_D1.4} to generate the MIMO channel matrix $\pmb{H}(t,n)$ for various parametrizations. 
This channel model has been validated for different propagation scenarios, as mentioned in \cite{METIS_D1.4}, and is a state-of-the-art channel model for next-generation wireless communication systems. 


The ray-tracer channel model considers a number of multipath components $k$ existing in the downlink communication between the BS-terminal pair, for each time $t$ and sub-carrier $n$. 
These multipath components arise due to different wave propagation phenomena, including reflection, diffraction and scattering, which are affected by the presence of scatterers in the propagation environment. 
The radiation patterns of the BS and terminal antennas are also taken into account by the ray-tracing model. 
We denote by $\tilde{h}_{k,a_{\textrm{u}},a_{\textrm{r}}}$ the impulse response for multipath component $k$, between each BS antenna element $a_{\textrm{r}}$ and each terminal's antenna element $a_{\textrm{u}}$, which captures all the aforementioned propagation effects in addition to the relevant pathloss. 
The channel impulse response $H_{a_{\textrm{u}},a_{\textrm{r}}}(t,n) \in \pmb{H}(t,n)$ is then the sum of the impulse responses of all the $k$ different multipath components, i.e.
\begin{align}
\label{eq:5:channel_model}
    H_{a_{\textrm{u}},a_{\textrm{r}}}(t,n) = \sum_{k = 1}^{K} \tilde{h}_{k,a_{\textrm{u}},a_{\textrm{r}}} \cdot \exp^{\frac{j2\pi d_k(t)}{\lambda}} \exp^{-j2\pi f_{n}\tau_{k,a_{\textrm{u}},a_{\textrm{r}}}(t)}. 
\end{align}

Here, $K$ is the total number of multipath components, $\lambda$ is the wavelength corresponding to the center frequency $f_\textrm{c}$, and $f_n$ is the frequency of the sub-carrier $n$. $d_k$ is the total distance for multipath $k$ at time $t$, and $\tau_{k,a_{\textrm{u}},a_{\textrm{r}}}$ denotes the delay for multipath $k$. 
A detailed implementation of this channel model can be found in \cite{Petteri_Beamforming_2016}.

Based on the scenario description and the choice of the channel model, we define the following baseline cases to generate the primary datasets:
\begin{itemize}
    \item Case 1: When no scatterers are present in the propagation environment and accurate position estimates for the terminal are available. This represents a deterministic channel generation, i.e. the channel between a given terminal position and BS always results in the same channel matrix $\pmb{H}(t, n)$.
    \item Case 2: When erroneous position estimates are available (specifically, $\sigma =$ 0.4 m) with no scatterers in the propagation scenario.
    \item Case 3: The position estimates of the terminal are known accurately for $\rho \leq $ 0.05/m$^2$ that are randomly placed in the propagation environment.
\end{itemize}

We now present the methodology for the formulation of datasets used in this work. 
Afterwards, we analyze the generated datasets for the three cases mentioned above.

\subsection{Formulation of the Datasets}
\label{subsec4:3:data_formulation} 

We use exhaustive search to determine the optimal resource allocation for each sample in the generated dataset. 
This implies that for every estimate of terminal's coordinates $\hat{\pmb{p}}(t)$, the exhaustive search is performed offline to determine the optimal $\mathfrak{r}(t)$. 
Depending on the propagation conditions, for the system model specified earlier, multiple resource allocations can yield the same transport capacity value, which is optimal for the BS-terminal pair during time $t$. 
As an example, this can occur when different placements of scatterers result in a range of transmission rates that lead to similar error probability for payload transmission.  
Therefore, for certain scenarios, a vector $\pmb{\mathfrak{r}}(t)$ results in an optimal $\mathcal{T}$ value at time $t$, instead of a scalar $\mathfrak{r}(t)$. 
This renders the dataset formulation to be unconventional compared to the formulations that are traditionally used for supervised learning. 
In this work, we formulate the datasets with single output variable per sample, where the following approaches are considered for dataset formulation: 
\begin{itemize}
    \item \textbf{Dataset 1 ($\pmb{D}_1$):} This is the approach we adopted in our previous work \cite{Sahar_GLOBECOM_2019}, where we consider only the first optimal solution for capacity maximization problem using exhaustive search to form the dataset. This means that the output variable is the first resource allocation outcome from exhaustive search that maximizes the transport capacity for a given position estimate. This formulation results in a dataset with unique input-output association. 
    \item \textbf{Dataset 2 ($\pmb{D}_2$):} Here, we consider only the resource allocation of the transport capacity maximization problem that relates to the highest value of $\gamma_{\textrm{eff}}$ for formulating the dataset. This also implies unique input-output relationship in the dataset used for a learning framework.
    \item \textbf{Dataset 3 ($\pmb{D}_3$):} In contrast to the above two approaches, we use all possible resource allocations that optimally solve problem \eqref{eq:4:RA_prob} for constructing this dataset. This means that the learning algorithm will have to predict a single resource allocation by learning on non-unique relationship between the position estimates and the resource allocations.  
\end{itemize}

Each of these datasets are generated with realistic system assumption, where an estimate of user position as well as its CSI is collected after certain time intervals over a long period of time. 
This assumption is implemented in simulation by considering random user drops, as mentioned earlier, with uniformly distributed user positions over the entire simulation scenario for a fixed system parametrization. 
For a given user position, the channel matrix is obtained through the ray-tracer channel model, which is then used to compute the SNR and the effective SNR values based on the three resource allocation variables, i.e. the transmit beam $\pmb{v}(t)$, the receive filter $\pmb{u}(t)$ and the MCS $m(t)$. 
For all the datasets, a dataset sample $\pmb{d}_i$ consists of an input vector and an output value. 
The input vector comprises the $x-$ and $y-$coordinates of the terminal's position estimate $\hat{\pmb{p}}(t)$ (the $z-$coordinate is not used due to the same receive antenna height assumption for all the samples).
The output value is a number denoting the binary sequence, where the sequence encodes the index of $\pmb{v}$, $\pmb{u}$ and $m$ corresponding to the resource allocation considered for the specific dataset formulation. 
Here, a resource allocation denotes a class to be learnt by the learning algorithm. 
A collection of such $\pmb{d}_i$ samples constitutes the whole dataset $\pmb{D}$, which is then divided into a training dataset $\pmb{D}'$ and a test dataset $\pmb{D}''$ for applying machine learning. 
Based on the dataset formulations outlined before, $\pmb{D}_1$ and $\pmb{D}_2$ have a unique input-output association per sample, for both the training and test datasets. 
For ${\pmb{D}'}_3$, the input vector is repeated as many times as the optimal resource allocations for a given $\hat{\pmb{p}}(t)$, to construct data samples with a single output value denoting each of the optimal resource allocations per position estimate. 
The dataset ${\pmb{D}''}_3$ has a structure similar to ${\pmb{D}''}_1$, with unique input-output association per sample. 

The above process is repeated multiple times by setting the system parametrization variables differently, to get various datasets for each formulation with the assumption of accurate user position availability.
For the case of erroneous position estimates, the inputs in the datasets for accurate user position are replaced by erroneous position estimates modelled by a zero-mean Gaussian with specific variance $\sigma^2$, while the outputs are kept the same as the ones in the accurate position datasets.

\subsection{Analysis of the Datasets}
\label{subsec4:4:dataset_analysis}

After generating the different dataset formulations, we analyze their input-output associations to have an intuition about the learning performance on a specific data formulation. 
A total of $I = $ 125,000 position estimates are generated for datasets $\pmb{D}_1$, $\pmb{D}_2$ and $\pmb{D}_3$, for each of the three cases mentioned in~\ref{subsec4:2:channel_model}. 
Two thirds of these position estimates are used for constructing training datasets $\pmb{D}'$ and their analysis is presented here, while the rest of the samples are used for test dataset construction. 

\begin{figure}[!h]
\centering
\includegraphics[width=8cm]{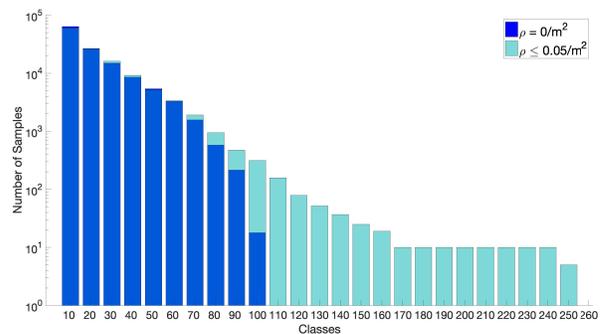}
\caption{Distribution of the number of samples per class in training dataset 1 ${\pmb{D}'}_1$, for case 1 ($\rho$ = 0/m$^2$) and case 3 ($\rho \leq$ 0.05/m$^2$), for $8 \times 2$ MIMO system.}
\label{fig:3:class_dist_d1}
\end{figure}

We start by presenting the number of samples per class (a class represents a unique resource allocation) distribution for the different dataset formulations. 
Note that the number of samples per class are the same for case 1 and 2, since they only differ in the input values. 
Fig.~\ref{fig:3:class_dist_d1} shows the distribution of the number of samples per class for ${\pmb{D}'}_1$, for case 1 and 3, where the x-axis is batched in groups of ten classes for better illustration. 
We observe an exponential distribution of the number of samples per class, which indicates that the learning can be influenced by a bias in favor of the dominantly occurring classes. 
We also observe that the number of classes for case 1 with deterministic channel generation (94) is almost $3\times$ lesser than that for case 3 (254), though being significantly smaller than the total number of generated data samples (125,000). 
The increased number of classes for case 3 is due to the fact that more scatterers result in more multipath components, which vary with scatterers' placement, introducing more randomness in the channel. 
This means that for a fixed user position with different placement of the scatterers, the channel response varies, and therefore, different RAs are obtained as the first outcome from the exhaustive search.

\begin{figure}[!h]
\centering
\includegraphics[width=8cm]{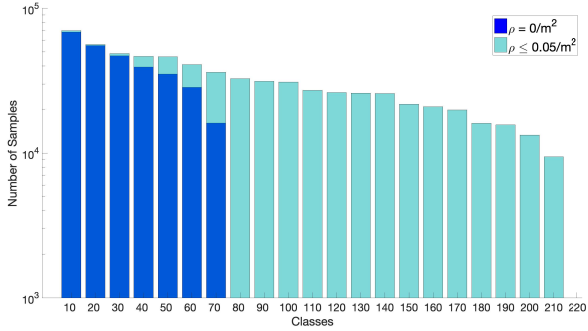}
\caption{Distribution of the number of samples per class in training dataset 2 ${\pmb{D}'}_2$, for (a) case 1 ($\rho$ = 0/m$^2$), and (b) case 3 ($\rho \leq$ 0.05/m$^2$), for $8 \times 2$ MIMO system.}
\label{fig:4:class_dist_d2}
\end{figure}

Fig.~\ref{fig:4:class_dist_d2} shows the distribution of number of samples per class for ${\pmb{D}'}_2$. 
Here again, the x-axis is batched in groups of ten classes for illustration purpose. 
In contrast to the observation for ${\pmb{D}'}_1$, an equitable distribution of the number of samples per class exists in ${\pmb{D}'}_2$. 
Although the number of classes for case 3 (209) is still $3\times$ that for case 1 (70), however, this is lesser than that observed in ${\pmb{D}'}_1$ due to the consideration of RA related to the highest effective SNR, which implies that the RA does not change significantly based on the random effects present in the wireless channel.

\begin{figure}[!h]
\centering
\includegraphics[width=8cm]{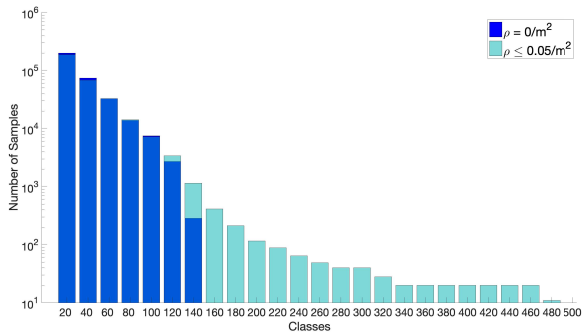}
\caption{Distribution of the number of samples per class in training dataset 3 ${\pmb{D}'}_3$, for (a) case 1 ($\rho$ = 0/m$^2$), and (b) case 3 ($\rho \leq$ 0.05/m$^2$), for $8 \times 2$ MIMO system.}
\label{fig:5:class_dist_d3}
\end{figure}

Fig.~\ref{fig:5:class_dist_d3} represents the number of samples distributed per class for ${\pmb{D}'}_3$. 
Note that the x-axis is batched in groups of 20 classes for better illustration. 
Here, we also observe an exponential distribution of the number of samples per class for both the cases, though the number of classes for both the cases in ${\pmb{D}'}_3$ is almost twice that for ${\pmb{D}'}_1$. 
This increase is a result of considering all RA outcomes from the exhaustive search that maximize the transport capacity for a given terminal position. 
The number of such outcomes varies based on the randomness in the propagation scenarios. 

\begin{figure}[!h]
\centering
\includegraphics[clip,trim={5cm 0cm 3cm 2cm},width=9cm]{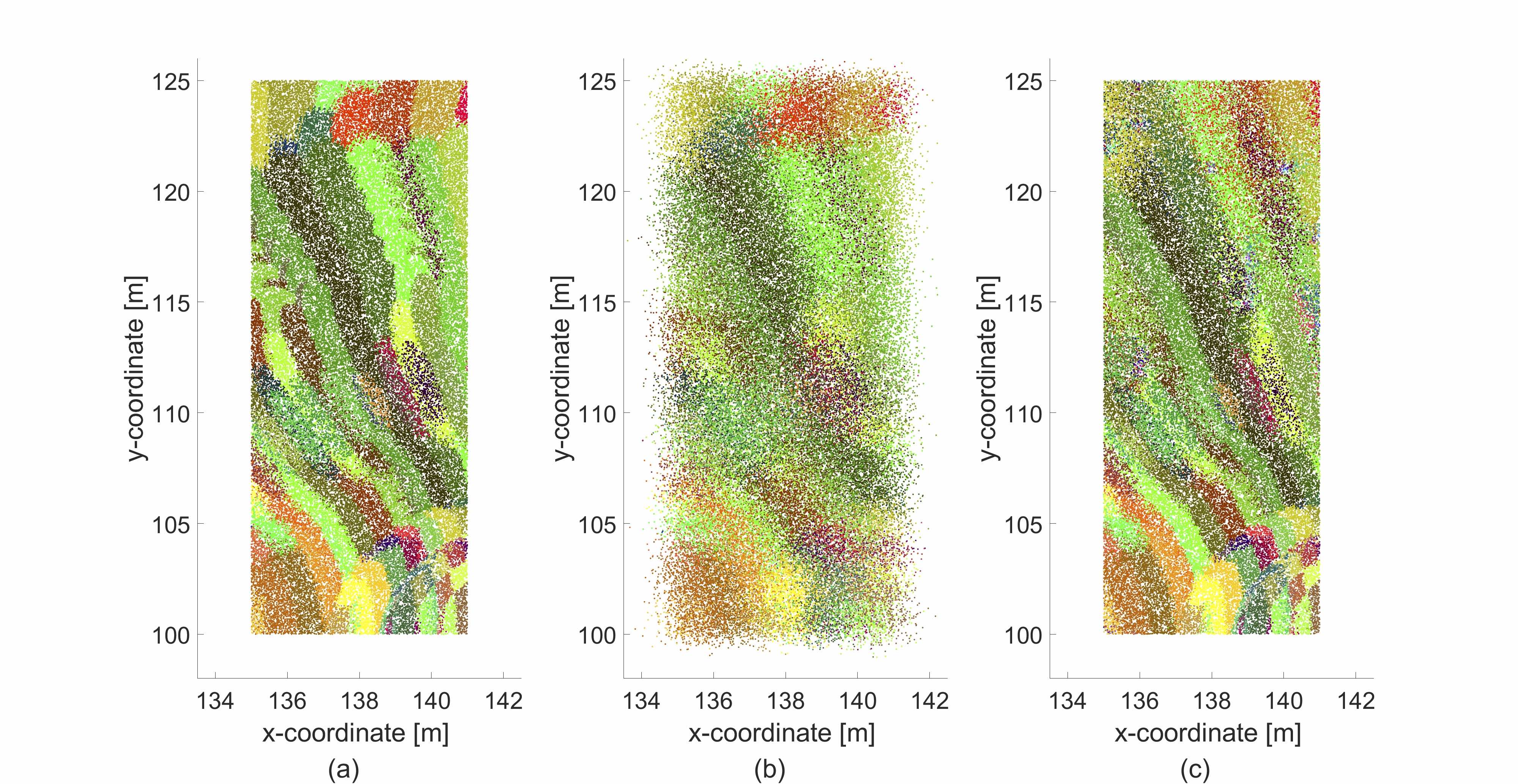}
\caption{Distribution of the classes with respect to user positions in ${\pmb{D}'}_1$, for (a) case 1: $\rho$ = 0/m$^2$, $\sigma =$ 0 m, (b) case 2: $\rho$ = 0/m$^2$, $\sigma = $ 0.4~m, and (c) case 3: $\rho \leq$ 0.05/m$^2$, $\sigma = $ 0 m, for $8 \times 2$ MIMO system.}
\label{fig:6:scatter_plot_d1}
\end{figure} 

After observing the distribution of inputs versus classes for the different dataset formulations, we now analyze the distribution of classes in relation to the input parameters. 
Fig.~\ref{fig:6:scatter_plot_d1} presents the distribution of classes, the resource allocations, in relation to the input parameters, the position coordinates of the terminal, for case 1, 2 and 3. 
Note that each resource allocation is depicted by a unique color, which is consistent across all the cases. 
For case 1, i.e. the deterministic channel, we see that the different classes are separated quite distinctly over most of the considered area. 
Some overlap between the classes occurs either in the middle part of the street section or in the lower right corner, which is closer to the BS. 
The former behavior of class overlap exists due to a higher MCS value maximizing the transport capacity within a specific distance range from the BS, while the latter behavior is a result of the antenna radiation pattern at the BS. 
Comparing Fig.~\ref{fig:6:scatter_plot_d1}(a) and (b), we observe that the class boundaries become dispersed as the position estimates become erroneous. 
This dispersion poses a challenge for the learning algorithms in terms of robustness. 
Considering the channel characteristics, we compare Fig.~\ref{fig:6:scatter_plot_d1}(a) and (c) and notice that the random effects in the propagation scenario do not impact the dataset as severely as the erroneous position estimates do. 
Though the number of classes in case 3 is thrice as that in case 1, the class boundaries are still distinctly separated across the street section. 
Therefore, the performance of the learning algorithm can be affected primarily by the erroneous position estimates for ${\pmb{D}'}_1$, compared to the presence of random scatterers in the propagation scenario. 

\begin{figure}[!h]
\centering
\includegraphics[clip,trim={5cm 0cm 3cm 2cm},width=9cm]{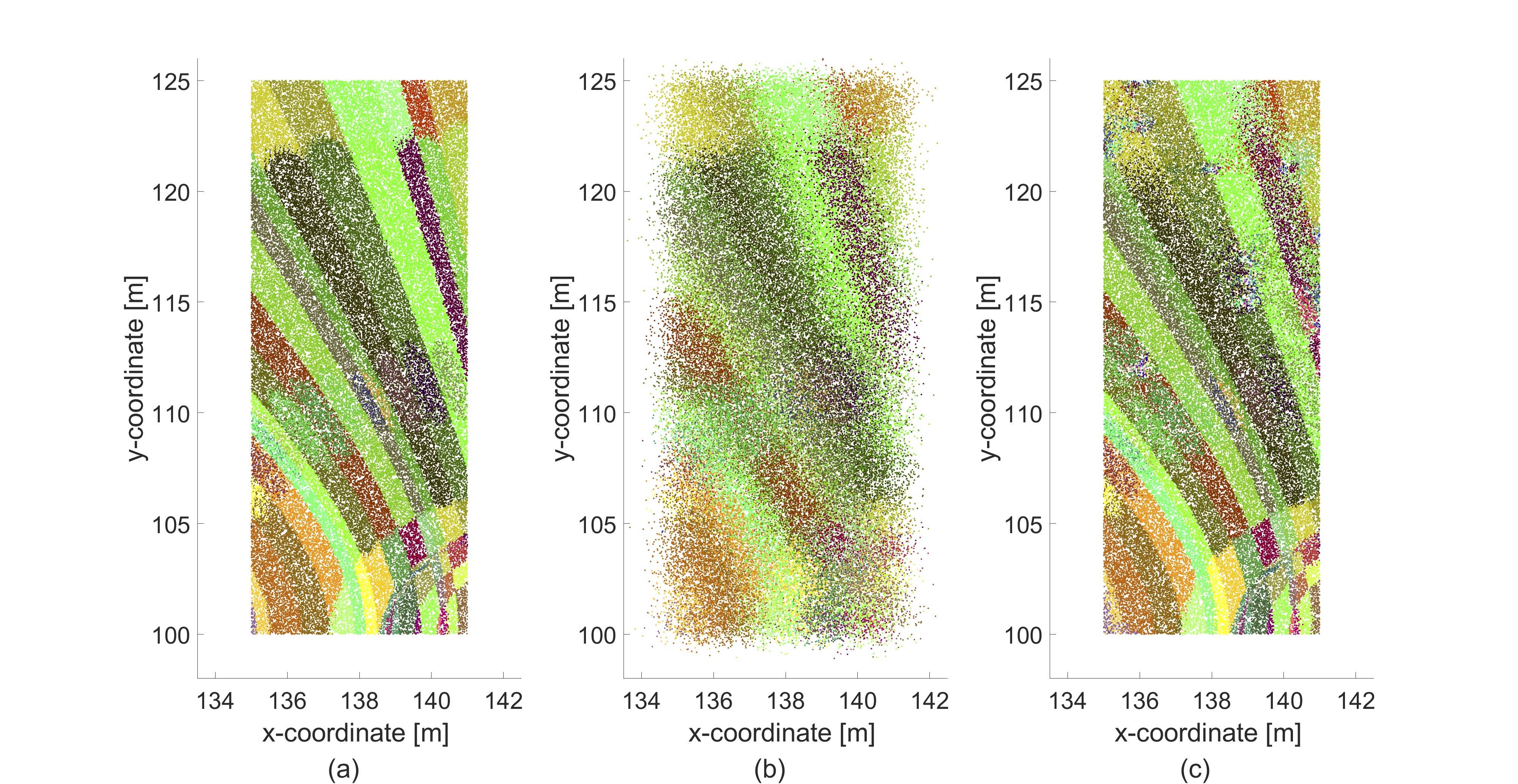}
\caption{Distribution of the classes with respect to user positions in ${\pmb{D}'}_2$, for (a) case 1: $\rho$ = 0/m$^2$, $\sigma =$ 0 m, (b) case 2: $\rho$ = 0/m$^2$, $\sigma = $ 0.4~m, and (c) case 3: $\rho \leq$ 0.05/m$^2$, $\sigma = $ 0 m, for $8 \times 2$ MIMO system.}
\label{fig:7:scatter_plot_d2}
\end{figure}

We now show the distribution of classes with respect to the terminal position estimates for $\pmb{D}_2'$ in Fig.~\ref{fig:7:scatter_plot_d2}. 
Note that we observe the same color coding as in Fig.~\ref{fig:6:scatter_plot_d1}, for better comparison across the two datasets ${\pmb{D}'}_1$ and ${\pmb{D}'}_2$. 
The most important observation from Fig.~\ref{fig:7:scatter_plot_d2} is that the classes are much distinctly separated, with clearly defined boundaries, for case 1. 
These class boundaries become dispersed due to the erroneous position estimates for case 2, however, the extent of dispersion remains the same as observed for ${\pmb{D}'}_1$, despite the clear class boundaries observed for case 1. 
The class boundaries for case 3 become blurred with the introduction of more classes, compared to case 1, but this blurriness is lesser than the one observed for case 2, i.e. for erroneous terminal positions.  

\begin{figure}[!h]
\centering
\includegraphics[clip,trim={5cm 0cm 3cm 2cm},width=9cm]{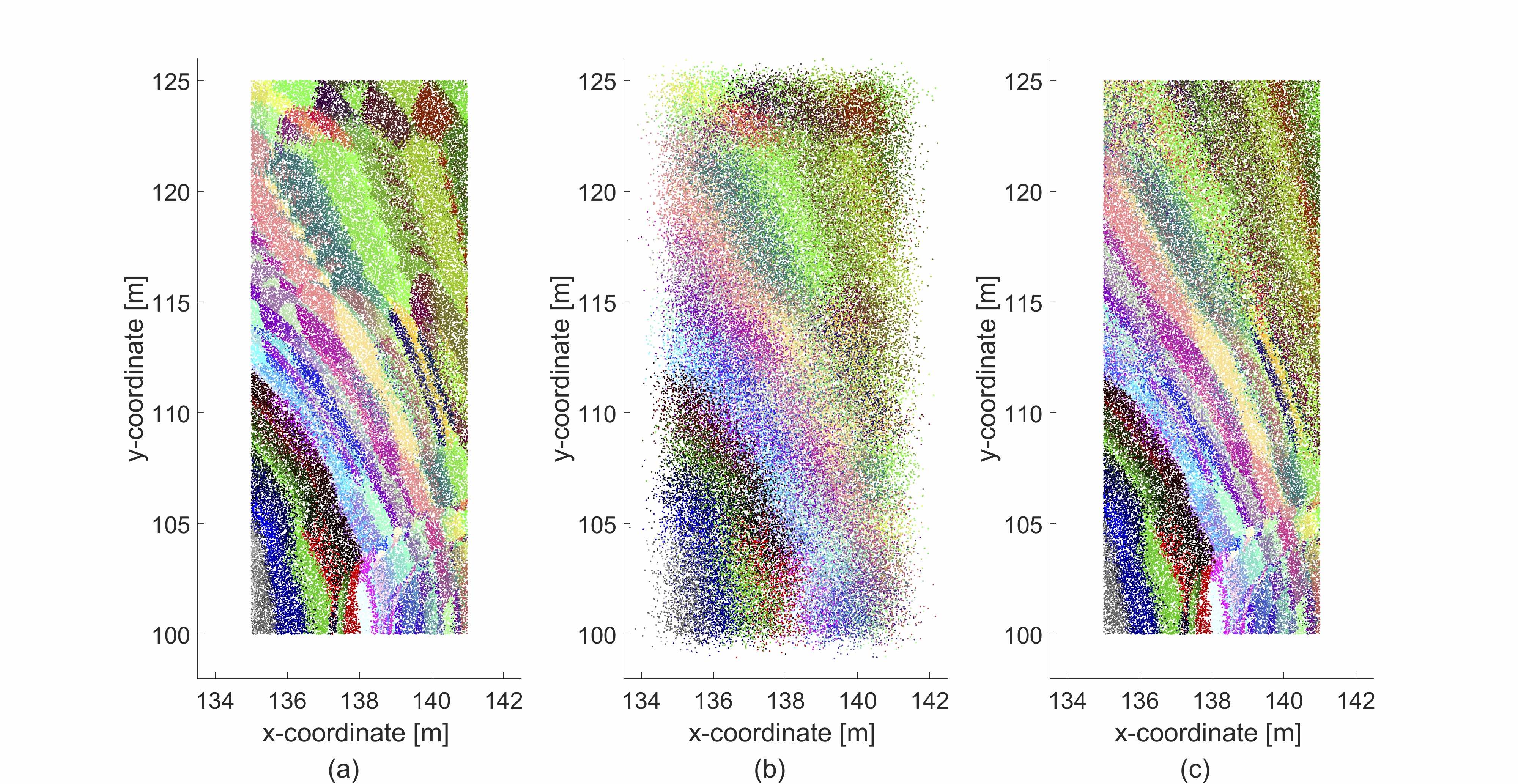}
\caption{Distribution of the classes with respect to user positions in ${\pmb{D}'}_3$, for (a) case 1: $\rho$ = 0/m$^2$, $\sigma =$ 0 m, (b) case 2: $\rho$ = 0/m$^2$, $\sigma = $ 0.4~m, and (c) case 3: $\rho \leq$ 0.05/m$^2$, $\sigma = $ 0 m, for $8 \times 2$ MIMO system.}
\label{fig:8:scatter_plot_d3}
\end{figure} 

The representation of distribution of classes with respect to the terminal positions for ${\pmb{D}'}_3$ is a complex task, since a single terminal position is associated with multiple resource allocations, or classes. 
One possible illustration is presented in Fig.~\ref{fig:8:scatter_plot_d3}, where a set of all resource allocations associated with a single terminal position is shown as a unique color. 
Based on this color coding, 1137 unique sets of resource allocations are identified in ${\pmb{D}'}_3$, and therefore, a color palette of 1137 colors is used for plotting Fig.~\ref{fig:8:scatter_plot_d3}. 
We observe the same behavior for case 1, 2 and 3, as observed previously for datasets ${\pmb{D}'}_1$ and ${\pmb{D}'}_2$, with one main difference: The blurriness or dispersion of a class boundary does not necessarily imply a performance loss in terms of transport capacity. 
This is due to the fact that a different color shows a unique set of RAs, with the possibility of certain RAs being common to both the unique sets. 
Because of this, we define a new criteria to determine the performance of the learning algorithms based purely on the dataset characteristic, which will be explained in the later sections. 
\begin{figure}[!h]
\centering
\includegraphics[width=6cm]{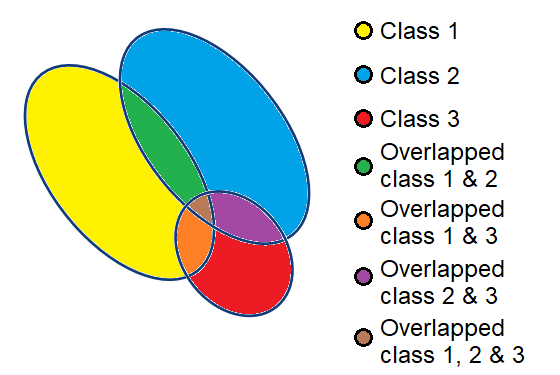}
\caption{Representative plot of the distribution of three classes with respect to user positions in ${\pmb{D}'}_3$ for $8 \times 2$ MIMO system.}
\label{fig:9:small_d3}
\end{figure}

An alternate representation of the distribution of classes in relation to the terminal positions for ${\pmb{D}'}_3$ is shown in Fig.~\ref{fig:9:small_d3}. 
Note that we only illustrate the distribution for three of the resource allocations, i.e. the classes, with respect to the terminal positions, to understand how the dataset ${\pmb{D}'}_3$ is viewed by the learning algorithm. 
The representation plot shows that the learning algorithm can associate the overlapping class regions with any of the classes for a given terminal position. 
This implies that the predicted class will result in the same value of maximum transport capacity, as long as the prediction lies within the original and overlapped class boundary. 
The non-unique association can assist the learning algorithm to be robust for erroneous position inputs, as the dispersed class boundaries will still be lying in the overlapped region and, therefore, result in lesser loss in transport capacity compared to that for datasets $\pmb{D}_1$ and $\pmb{D}_2$. 

In this section we presented the different dataset formulations for supervised machine learning to perform coordinates-based resource allocation. 
Analysis of the different dataset formulations shows that strong spatial clustering exists, which supports the feasibility of coordinates-based resource allocation through supervised machine learning frameworks. 
The learning task can be challenging for some dataset formulations, for specific propagation scenarios, which yields interesting results, as discussed in the next section of this paper. 

\section{Evaluation Results and Discussion}
\label{sec:5:results_main_section} 
In this work we are interested in evaluating the applicability of coordinates-based RA under different propagation scenarios and system constraints, as well as the computational resources for implementing the proposed scheme in a realistic system setup. 
Based on the scenario description in Section~\ref{subsec4:4:dataset_analysis}, the performance of the following schemes is evaluated in our work:
\begin{itemize}
    \item \textbf{CSI-based RA scheme:} This represents the traditionally used RA scheme which relies on the instantaneous CSI of the BS-terminal pair. 
    \item \textbf{Coordinates-based RA scheme using RF:} As discussed in Section~\ref{subsec3:3:ML}, this scheme uses the random forest algorithm for learning the training dataset. The details for selecting the different parameters of RF model are given in Section~\ref{subsec5:2:tuning_RF}. 
    \item \textbf{Coordinates-based RA scheme using KNN:} As discussed in Section~\ref{subsec3:3:ML}, this is the simplest machine learning-based scheme, where we consider $K = 1$. 
    \item \textbf{Geometric-based RA scheme:} This is the benchmark scheme, where geometric beamforming is used to determine the transmit beam and receiver filter based on the terminal's coordinates in relation to the BS placement. The MCS is determined statistically based on the terminal's position coordinates in relation to the geometry of propagation scenario.
\end{itemize} 

The above performance evaluation is done for different channel characterizations: When no scatterers are present in the propagation environment, or when a number of scatterers up to 5 per 100 m$^2$ are randomly placed in the propagation environment, and accurate position estimates for the terminal are available to the system. 
Recall that the former case represents a deterministic channel generation (as mentioned in \ref{subsec4:2:channel_model}), while the latter refers to a varying channel generation case. 
The performance comparison for different data formulations is also done for a fairly deterministic channel generation, where the scatterer density $\rho$ varies up to 1 scatterer per 100 m$^2$. 
To investigate the performance limit of the proposed coordinates-based RA scheme, we consider different degrees of variation in the error associated with the estimated coordinates of the terminal. 
This variation is defined by $\sigma~=~0, 0.25, 0.4$ and $1$ m. 
We now we discuss the tuning of RF algorithm for generating the results related to various propagation scenarios considered in our work, followed by the performance results and relevant discussions. 
At the end, we discuss the implementation of coordinates-based RA scheme in a realistic system setup and also comment on the computational resources needed for its implementation. 


\subsection{Tuning of the Random Forest Algorithm}
\label{subsec5:2:tuning_RF}

\begin{figure}[!h]
\centering
\includegraphics[width=8cm]{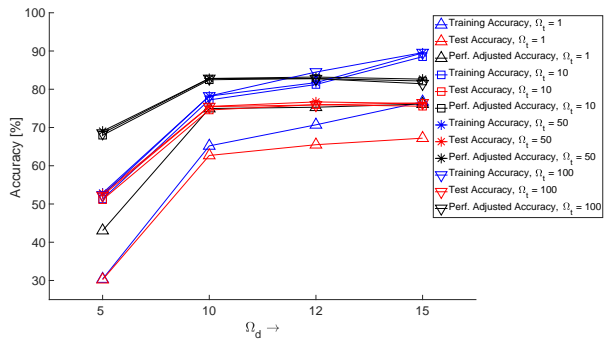}
\caption{Accuracy results for different parametrization of Random Forest.}
\label{fig:10:result_RF_tuning}
\end{figure}

To optimize the performance of random forest algorithm, we need to tune its parameters. 
Specifically, we need to decide on the number of trees $\Omega_{\textrm{t}}$ that make up the forest as well as the maximum depth $\Omega_{\textrm{d}}$ up to which each tree has to be built while training the model. 
In terms of the number of randomly selected input features $\pmb{\mathfrak{f}}'$ for building each node of the tree, we resort to the conventional practice, i.e. we choose $\pmb{\mathfrak{f}}' = \sqrt{\pmb{\mathfrak{f}}}$. 
Fig.~\ref{fig:10:result_RF_tuning} shows the training and test accuracy obtained for different number of trees, at varying maximum depth per tree, for the RF algorithm. 
These tuning results are shown for the maximum scatterers' density, i.e. $\sigma \leq$ 0.05/m$^2$, with data formulation $\pmb{D}_2$. 
In addition to the traditionally used training and test accuracy metrics, we define a new accuracy metric to determine the performance of the learning algorithm in comparison to the throughput maximization problem. 
Generally, the test accuracy is computed using one-to-one comparison between the ground truth label and the predicted label in the test dataset. 
However in our work, as mentioned before, each sample of the dataset can have multiple outputs (the RAs) as ground truth labels, therefore, the test accuracy metric alone can not determine how well the learning algorithm has been trained. 
We call the newly defined metric as \emph{performance adjusted} accuracy, which is determined by comparing the predicted label to all possible set of labels associated with a test sample, instead of only a single label. 
Comparing the three accuracy metrics in Fig.~\ref{fig:10:result_RF_tuning}, we observe that the depth of trees has the most impact on the learning performance. 
A shallow depth, such as $\Omega_{\textrm{d}} = 5$, for the trees is not sufficient to learn the relationship between $\hat{\pmb{p}}(t)$ and $\pmb{\mathfrak{r}}(t)$, but the depth can be increased up to a certain extent, such as $\Omega_{\textrm{d}} = 15$ to prevent the model to over-fit the training dataset. 
The test accuracy increases till $\Omega_{\textrm{d}} = 15$, while the performance adjusted accuracy remains fairly constant. 
In addition, the performance does not seem to be affected by the number of trees, as long as it is sufficiently high. 
Furthermore, a higher number of trees introduces variance in the learnt model, which means that the trained model can learn the various classes even with smaller number of the associated training samples. 
Note that even for higher number of trees the model requires only a little more training time and memory to be stored by the system. 
Based on the above observations, we choose the RF parametrization to be $(\Omega_{\textrm{t}}, \Omega_{\textrm{d}}) = (100, 15)$.

\begin{figure*}
\centering
\includegraphics[width=18cm]{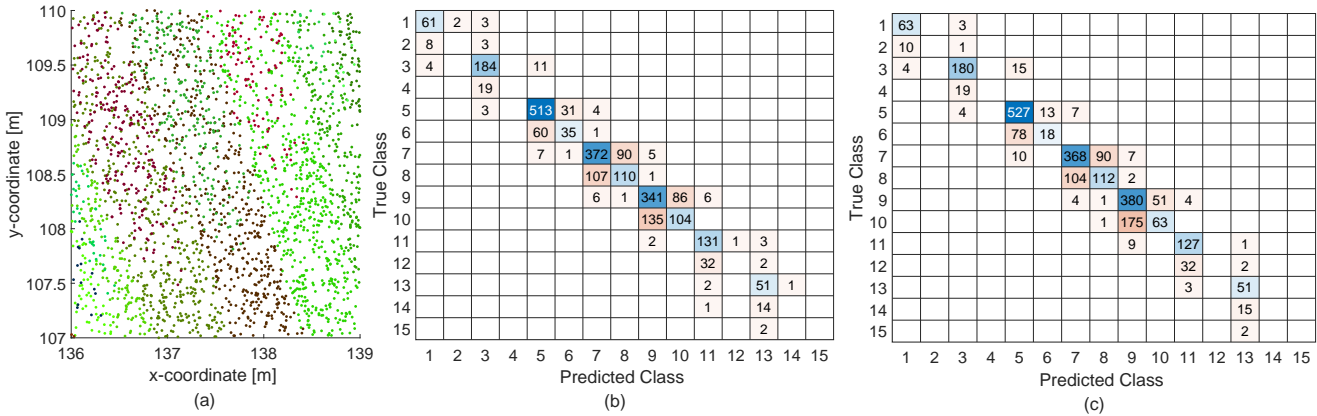}
\caption{(a) Distribution of classes with respect to user positions for a small area of the street-section, with test dataset 2 for $\rho \leq 0.05$/m$^2$ in $8\times2$ MIMO system, (b) Confusion matrix for the small street section in (a) with $\Omega_t = 100$, $\Omega_d = 15$ for Random Forests algorithm, and (c) Confusion matrix for the small street section in (a) with $\Omega_t = 50$, $\Omega_d = 12$ for Random Forests algorithm..}
\label{fig:11:conf_mat_scatter_plot}
\end{figure*}

For better understanding of the RF model, we focus on the confusion matrix obtained for a part of the street section shown in Fig.~\ref{fig:7:scatter_plot_d2}(c). 
We focus on two parametrizations of RF that provide the best results according to Fig.~\ref{fig:10:result_RF_tuning}: $(\Omega_{\textrm{t}}, \Omega_{\textrm{d}}) = (100, 15)$ and $(50,12)$. 
The confusion matrix is a tabulated summary of the performance of classifier; each entry in the row of confusion matrix shows how many samples for the true class are confused with one of the predicted classes.  
Fig.~\ref{fig:11:conf_mat_scatter_plot}(a) shows the street section with 31 unique classes, or RAs, while Fig.~\ref{fig:11:conf_mat_scatter_plot}(b) and (c) show the confusion matrices for that street-section obtained for the aforementioned settings of RF algorithm. 
Comparing the confusion matrices, we see that RF with $(\Omega_{\textrm{t}}, \Omega_{\textrm{d}}) = (100, 15)$ shows equivalent classification rate for all the classes (marked on the diagonal) compared to $(\Omega_{\textrm{t}}, \Omega_{\textrm{d}}) = (50, 12)$. 
This confirms our previous observation and therefore, the parametrization of $(\Omega_{\textrm{t}}, \Omega_{\textrm{d}}) = (100, 15)$ is used for evaluating the proposed RA scheme. 

\begin{figure}[!h]
\centering
\includegraphics[width=8cm]{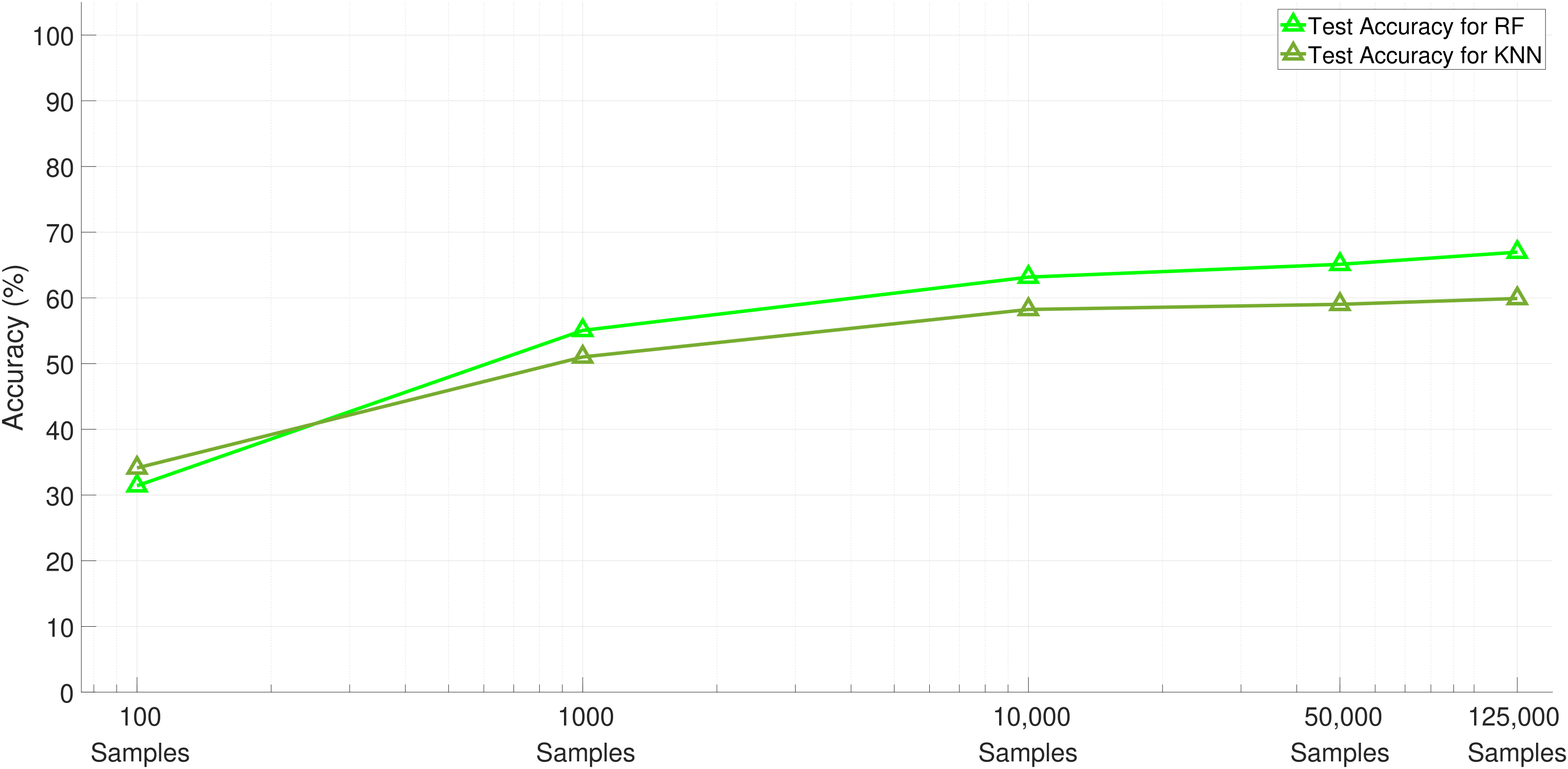}
\caption{Variation in training and test accuracy for different number of samples in the dataset.}
\label{fig:12:result_RF_sample_var}
\end{figure}

Another important factor to consider while training a learning model is to determine the number of training samples required to achieve a reasonable performance. 
Fig.~\ref{fig:12:result_RF_sample_var} shows the test accuracy for KNN and RF algorithm for different number of samples in the overall dataset. 
The results show that the test accuracy saturates for 10,000 samples in the dataset, for both the learning frameworks. 
With very small number of samples, KNN performs better than RF, but as the number of samples increase, the RF performs consistently well compared to KNN. 
The advantage of random selection with replacement is not beneficial for RF when very small number of samples are available, but with increased number of samples, RF can achieve up to 10\% better accuracy than the simplest learning framework, the KNN. 
Overall, the test accuracy is the highest for a total of 125,000 samples in the dataset, and that is why, we evaluate the performance of all the schemes for a dataset size of 125,000 samples in the next sub-section. 

\subsection{Performance Results and Discussion}
\label{subsec5:3:all_results}

\begin{figure}[!h]
\centering
\includegraphics[clip,trim={0 1cm 3cm 0},width=8cm]{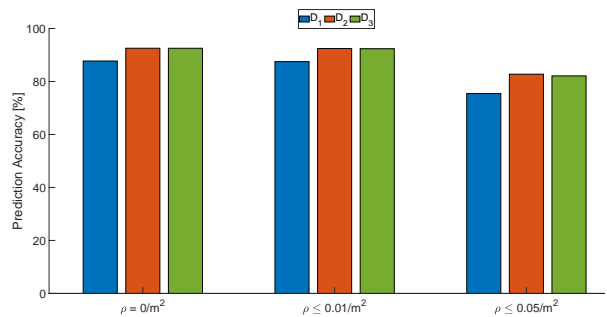}
\caption{Prediction accuracy for different dataset formulations, for 8$\times$2 MIMO system with accurate position estimates for different scatterers' densities.}
\label{fig:13:result1_1}
\end{figure}

First, we present the results related to the considered dataset formulation.  
Fig.~\ref{fig:13:result1_1} shows the performance adjusted accuracy evaluated for test datasets ${\pmb{D}''}_1$, ${\pmb{D}''}_2$ and ${\pmb{D}''}_3$, with $8\times2$ MIMO system for various scatterers' densities when accurate position estimates are available to the system. 
We observe that ${\pmb{D}''}_2$ and ${\pmb{D}''}_3$ show comparable performance adjusted accuracy, irrespective of the number of scatterers present in the propagation environment. 
Performance adjusted accuracy for ${\pmb{D}''}_1$ is lower due to the fact that the number of samples per class distribution is highly exponential, as shown in Section~\ref{subsec4:4:dataset_analysis}: Fig.~\ref{fig:3:class_dist_d1}, and thus the samples belonging to less frequently occurring classes are misclassified most of the time. 
For training dataset ${\pmb{D}'}_2$, the samples per class distribution shows uniform behavior, whereas for ${\pmb{D}'}_3$, the learning framework learns on all possible $\pmb{\mathfrak{r}}(t)$ for each $\hat{\pmb{p}}(t)$, and therefore, both dataset formulations are less susceptible to misclassification. 
Table~\ref{table:1:capacity_all_schemes} shows the average transport capacity for the different RA schemes for different dataset forumulations. 
Overall, the dataset formulation does not affect the performance of KNN-based and RF-based RA schemes, irrespective of the considered propagation scenario. 
In terms of performance comparison between different schemes, the proposed coordinates-based resource allocation scheme achieves a transport capacity very close to the upper bound, the CSI-based RA scheme, which is twice as much as the one achieved by the benchmark geometry-based RA scheme. 
The geometry-based RA scheme relies on only the position estimate of the terminal to determine the resource allocation, disregarding the presence of scatterers in the propagation environment, and thus suffers from deteriorated performance. 
In general, both $\pmb{D}_2$ and $\pmb{D}_3$ show similar performance with respect to average transport capacity metric, but due to the ease of analysis of $\pmb{D}_2$, as discussed in Section~\ref{subsec4:4:dataset_analysis}, we will use $\pmb{D}_2$ for performance evaluation in the rest of the paper. 

\begin{table*}[!t]
\caption{Average System Transport Capacity for $8\times2$ MIMO system, for different scatterers' densities, with accurate position information}
\label{table:1:capacity_all_schemes}
\centering
\begin{tabular}{cccc}
\specialrule{.2em}{.1em}{.1em} 
Dataset Formulation & RA Scheme & $\rho =$ 0/m$^2$ & $\rho \leq$ 0.05/m$^2$\\
\specialrule{.2em}{.1em}{.1em} 
$\pmb{D}_1$, $\pmb{D}_2$, $\pmb{D}_3$ & CSI-based & $1.2082\times10^8$ [bps] & $1.2096\times10^8$ [bps] \\ \specialrule{.2em}{.1em}{.1em}
\multirow{3}{*}{$\pmb{D}_1$} & RF-based & $1.189\times10^8$ [bps] & $1.1382\times10^8$ [bps] \\\cline{2-4}
                                                     & KNN-based & $1.1842\times10^8$ [bps] & $1.11\times10^8$ [bps] \\\cline{2-4}
                                                     & Geometry-based & $0.7058\times10^8$ [bps] & $0.6656\times10^8$ [bps] \\
                                                     \specialrule{.2em}{.1em}{.1em} 
\multirow{3}{*}{$\pmb{D}_2$} & RF-based & $1.1908\times10^8$ [bps] & $1.1483\times10^8$ [bps] \\\cline{2-4}
                                                     & KNN-based & $1.1856\times10^8$ [bps] & $1.113\times10^8$ [bps] \\\cline{2-4}
                                                     & Geometry-based & $0.7035\times10^8$ [bps] & $0.6626\times10^8$ [bps] \\
                                                     \specialrule{.2em}{.1em}{.1em} 
\multirow{3}{*}{$\pmb{D}_3$} & RF-based & $1.1883\times10^8$ [bps] & $1.1488\times10^8$ [bps] \\\cline{2-4}
                                                     & KNN-based & $1.1828\times10^8$ [bps] & $1.1077\times10^8$ [bps] \\\cline{2-4}
                                                     & Geometry-based & $0.7035\times10^8$ [bps] & $0.6626\times10^8$ [bps] \\
                                                     \specialrule{.2em}{.1em}{.1em} 
\end{tabular}
\end{table*}

\begin{figure}[!h]
\centering
\includegraphics[clip,trim={0 1cm 3cm 0},width=8cm]{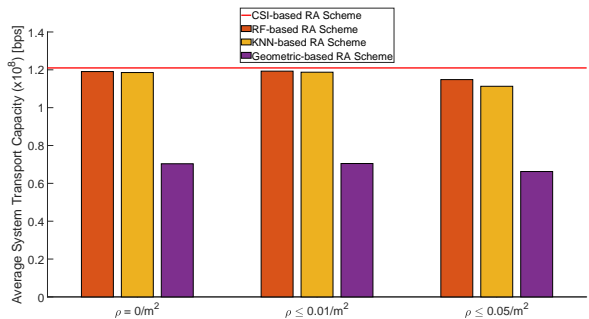}
\caption{Average transport capacity for 8$\times$2 MIMO system, with accurate position estimates for different density of scatterers.}
\label{fig:14:result2}
\end{figure}

Fig.~\ref{fig:14:result2} shows the average transport capacity of the system when the density of scatterers varies in the propagation environment, with the assumption that the position estimates of the terminals are accurately known to the system. 
The results show that both the KNN- and RF-based RA schemes are robust to the variation in scatterers' density compared to the CSI-based RA scheme, where a higher scatterers' density of $\rho \leq$ 0.05/m$^2$ leads to a performance difference of about 5\% compared to $\rho =~$0/m$^2$, when the channel is deterministic. 
RF-based RA scheme performs better than KNN-based scheme due to the inherent randomness in the random trees that constitute the RF model, and thus can cater for the random channel behavior due to randomized scatterers' placement, typically for the case of $\rho \leq$ 0.05/m$^2$. 
We conclude from the above discussion that the coordinates-based resource allocation scheme using machine learning can be applied for determining appropriate resource allocations under favorable propagation scenarios, without relying on CSI-collection. 
We now discuss the impact on the performance of the proposed scheme when either different antenna configurations are considered or when the erroneous position estimates are available in the system. 

\begin{figure}[!h]
\centering
\includegraphics[clip,trim={0 2cm 3cm 0},width=8cm]{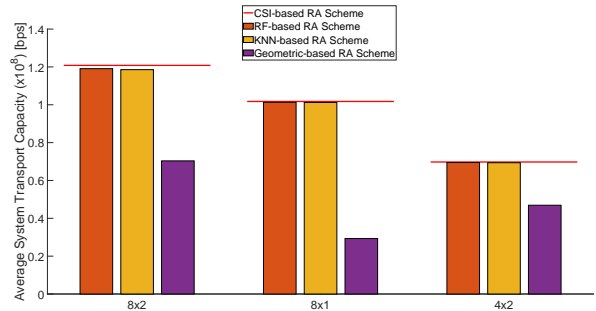}
\caption{Average transport capacity for different antenna configurations, with accurate position estimates for $\rho = $ 0/m$^2$.}
\label{fig:15:result3_1}
\end{figure}
 
\begin{figure}[!h]
\centering
\includegraphics[clip,trim={0 2cm 3cm 0},width=8cm]{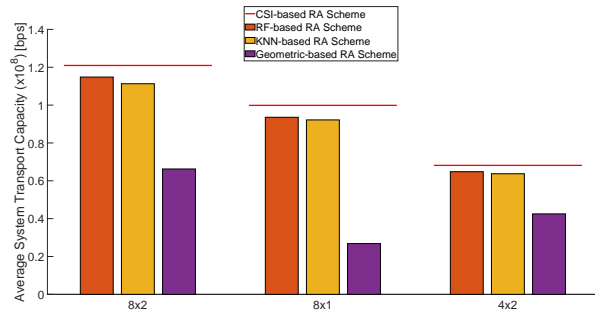}
\caption{Average transport capacity for different antenna configurations, with accurate position estimates for $\rho~\leq~$0.05/m$^2$.}
\label{fig:16:result3_2}
\end{figure}

Fig.~\ref{fig:15:result3_1} shows the average transport capacity for different antenna configurations when no scatterers are present in the propagation environment, while Fig.~\ref{fig:16:result3_2} shows the same when the scatterer density varies up to 0.05/m$^2$. 
The first observation is that the average system capacity drops by 50\% when the number of transmit antennas are reduced by half, due to the wider beam pattern which is a consequence of reduced number of antennas. 
The transport capacity decreases also when the number of receive antennas is reduced to one, since no receive beamforming can be applied to enhance the received power for a given position of the terminal. 
The performance of the proposed scheme, however, is not affected by the antenna configuration in general. 
For the case with no scatterers, as shown in Fig.~\ref{fig:15:result3_1}, the coordinates-based RA scheme performs on par with the CSI-based scheme for all antenna configurations, whereas the scheme performs consistently well when the scatterers' density varies for any antenna configuration (Fig.~\ref{fig:16:result3_2}). 
These results indicate that the proposed scheme can be used reliably with any antenna configuration for favorable propagation environments.

\begin{figure}[!h]
\centering
\includegraphics[width=8cm]{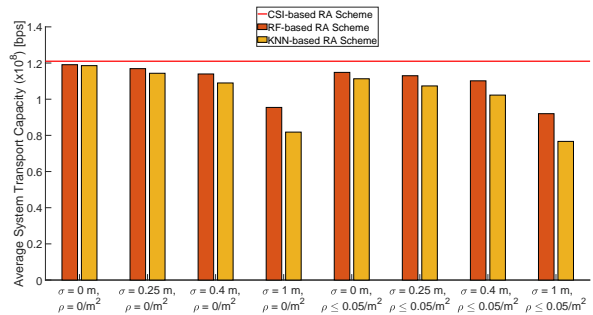}
\caption{Average transport capacity for 8$\times$2 MIMO system with different position inaccuracies for $\rho = $~0/m$^2$, and $\rho~\leq~$0.05/m$^2$.}
\label{fig:17:result4_1}
\end{figure} 

Another important aspect of investigation relates to the accuracy of the acquired position estimates of the terminals. 
Fig.~\ref{fig:17:result4_1} shows the average transport capacity of the system when the position estimates are known with varying error margins, for both the deterministic channel case as well as the randomly varying channel. 
RF-based RA scheme is very robust to the different degrees of error in the acquired position estimates, compared to KNN-based RA scheme, because of the inherent randomness in the trained RF model. 
A performance loss of about one-fifth of transport capacity is observed when the acquired position estimates are highly erroneous, i.e. $\sigma = 1$ m, for RF-based RA scheme, but is consistent for an error $\sigma \leq 0.5$ m. 
KNN-based RA scheme also performs consistently for position estimates having an error up to $\sigma \leq 0.5$ m, but results in a performance loss of one-third of the transport capacity compared to the CSI-based scheme when $\sigma = 1$ m. 
In general, the coordinates-based resource allocation scheme using machine learning performs at par with the legacy CSI-based scheme and is robust to the randomness introduced by the presence of scatterers for favorable propagation environment. 
The proposed scheme is also not affected by the considered antenna configuration and is quite robust to the erroneous position estimates acquired by the system, unless the position estimates are highly erroneous. 
All these observations are based on a sizeable amount of data acquired for uncorrelated samples. 
Next, we discuss the performance of the proposed RA scheme in comparison to other schemes when the dataset is constructed assuming real-time system simulation.

\subsection{Performance Evaluation for Correlated Channels}
\label{subsec5:2:correlated_channel}

After observing the feasibility of the coordinates-based RA scheme through machine learning on the datasets comprising uncorrelated samples, we now evaluate its performance on a realistic-system implementation. 
In real time, the data samples collected during the training-based mode of the proposed RA scheme (see Fig.~\ref{fig:1:scheme}) are collected on a continual basis, i.e. the collected samples have correlated channels associated with the estimates of the terminal's position $\hat{\pmb{p}}(t)$. 
These samples are then used to train the machine learning model, which is used for predicting the RA for a newly available position estimate to the system during the operation in position-based mode. 
For realistic-system implementation of the proposed coordinates-based RA scheme, the following key questions arise: (a) How many number of samples are sufficient to train a learning model, (b) how much training time is need to build the RA prediction model, and (c) how much time does the model take to predict a RA for a new $\hat{\pmb{p}}(t)$? 
In this work, we try to provide an intuition to answer these questions by designing a simple set of experiments. 

We resort to simulation-based set up, with the key idea of selecting a channel model that captures realistic channel behavior as accurately as possible. 
As mentioned before, the ray-tracer based METIS channel model~\cite{METIS_D1.4} has been validated for different propagation scenarios and is the state-of-the-art channel model available to date. 
Therefore, we use this channel model for the small street section, with the same system parametrization as considered in all the other experiments. 
Instead of using random-drop, the terminal moves in a straight line across the street so that the collected samples have correlated channel. 
The starting position of the terminal is generated randomly, and the subsequent samples are collected by updating only the $y-$coordinates of the terminal's position. 
We call the movement of the terminal along the street as a trace, and collect several traces for generating the dataset. 
Each sample is collected after a time period of 1 ms in a single trace, and a collection of these samples is then used to construct the training dataset according to the dataset formulation $\pmb{D}_2$. 
For a realistic system implementation, we assume the scatterers' density to be $\sigma \leq$ 0.05/m$^2$, where the number of scatterers as well as their placement varies independently over each trace.  
Overall, 50 traces were generated to have the training dataset size comparable to that of the uncorrelated dataset, for fair evaluation. 
In terms of the RF parametrization, we use the RF model with $(\Omega_{\textrm{t}}, \Omega_{\textrm{d}}) = (50, 12)$ for better real-time performance. 
To evaluate the performance of the trained RF model, we use the test dataset $\pmb{D}_2''$ to emulate the real-time data acquisition when the system operates in the position-based mode. 
We compute the average system transport capacity for the proposed coordinates-based RA scheme with KNN and RF models, and compare it to the one obtained for the CSI-based scheme. 

\begin{figure}[h!]
\centering
\includegraphics[width=8cm]{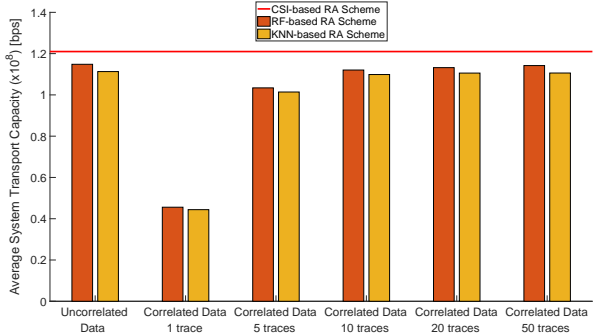}
\caption{System performance for 8$\times$2 MIMO system with uncorrelated and correlated channel datasets for $\rho~\leq~$0.05/m$^2$ with perfect position information. The number of traces in correlated channel dataset varies from 1 to 50.}
\label{fig:18:result_correlated_perfect_pos}
\end{figure} 

Fig.~\ref{fig:18:result_correlated_perfect_pos} shows the performance results for the three RA schemes when either uncorrelated samples are used for training or when different number of traces in the correlated dataset are used for training the machine learning models. 
We observe that a small number of traces are sufficient for generating the training dataset to achieve a performance comparable to the CSI-based RA scheme. 
Specifically, training dataset size of $\sim$10,000 samples, collected over a period of about 17 seconds, is capable of achieving a performance very close to the upper-bound, the CSI-based RA scheme. 
Furthermore, this performance is achieved with an RF model that needs only 937 bytes of memory storage for predicting the resource allocation $\mathfrak{r}(t)$. 
This is a surprising result: In real-time, the proposed coordinates-based scheme needs only a couple of seconds to collect the training data, and provides a system performance very close to the CSI-based scheme with only a small-sized learnt model. 

\begin{figure}[!h]
\centering
\includegraphics[width=8cm]{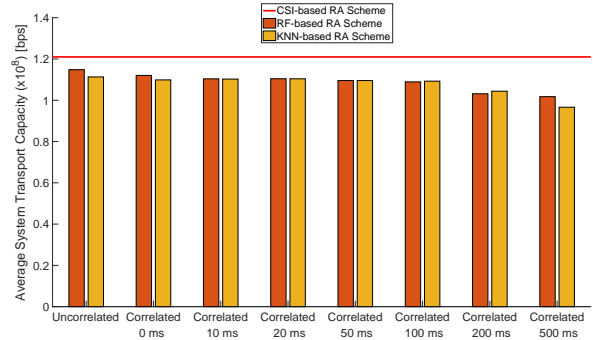}
\caption{System performance for 8$\times$2 MIMO system with uncorrelated dataset and correlated  dataset having 10 traces for $\rho~\leq~$0.05/m$^2$ with perfect position information. The undersampling in the correlated channel dataset  varies from 0 to 500 ms.}
\label{fig:19:result_correlated_perfect_pos_undersample}
\end{figure}

Efficient data collection process is vital for implementing the proposed coordinates-based RA scheme. 
This means that we also need to determine how frequently the training samples need to be acquired by the system. 
We apply different rates of undersampling on the correlated dataset with 10 traces, to see how stable is the performance when fewer number of samples are available to train the machine learning frameworks. 
Fig.~\ref{fig:19:result_correlated_perfect_pos_undersample} shows the average system transport capacity obtained on the uncorrelated test samples when the machine learning frameworks are trained with datasets of decreasing sample size. 
The results show that both the RF-based and KNN-based RA schemes have stable performance, unless extreme rate of undersampling is applied. 
An important observation here is that the performance of RF is at par with KNN even when the undersampling rate is as small as 10 ms. 
To analyze this performance variation between RF and KNN as the number of training samples decreases, we look into the difference of the predicted performance between the two machine learning frameworks. 
Table~\ref{table:2:RF_KNN_performance_diff} mentions the performance difference between RF and KNN averaged over the number of samples where the chosen machine learning framework outperforms the other. The results show that the margin by which one machine learning framework outperforms the other is consistent across the different datasets, and the margin decreases as the number of samples in the training dataset becomes small. 

\begin{table*}[]
    \centering
    \caption{Goodput Performance Difference (in [bps]) Between RF and KNN for Different Training Datasets}
    \begin{tabular}{|c|c|c|c|c|}
    \hline
         Samples where: & Uncorrelated Data & Correlated Data (50 traces) & Correlated Data (10 traces) & Correlated Data (10 traces, 10 ms)  \\
         \hline
         RF is better & $5.1912\times10^7$ & $5.1091\times10^7$ & $5.0517\times10^7$ & $4.3956\times10^7$ \\
         KNN is better & $3.3231\times10^7$ & $3.127\times10^7$ & $3.4437\times10^7$ & $3.2056\times10^7$ \\
         \hline
    \end{tabular}
    \label{table:2:RF_KNN_performance_diff}
\end{table*}

We also measured the impact of the performance difference between RF and KNN across different datasets by taking the sum of the transport capacity for the samples when one of the two machine learning frameworks performs better and normalize this sum by the total number of samples in the test dataset, i.e. 41,667. 
Table \ref{table:3:RF_KNN_normalized_perf} shows the resulting values, which indicate that RF outperforms KNN significantly when the number of training samples is sufficiently large (of the order of 10,000), but reduces drastically when a small number of training samples is available. 
Essentially, KNN performs at par with RF because the latter loses the generalization property due to lack of sufficient data for accurate classification.
    
\begin{table*}[]
    \centering
    \caption{Normalized Performance Difference (in [bps]) Between RF and KNN for Different Training Datasets}
    \begin{tabular}{|c|c|c|c|c|}
    \hline
         Samples where: & Uncorrelated Data & Correlated Data (50 traces) & Correlated Data (10 traces) & Correlated Data (10 traces, 10 ms)  \\
         \hline
         RF is better & $13.375\times10^6$ & $11.15\times10^6$ & $7.0833\times10^6$ & $5.7831\times10^5$ \\
         KNN is better & $6.2642\times10^6$ & $4.8168\times10^6$ & $2.8122\times10^6$ & $3.3929\times10^5$ \\
         \hline
    \end{tabular}
    \label{table:3:RF_KNN_normalized_perf}
\end{table*}

\begin{table*}[]
    \centering
    \caption{Training Time and Per-sample Prediction Time for RF Model, for Different Sizes of Real-time Dataset}
    \begin{tabular}{|c|c|c|c|c|}
    \hline
         Parameters & 50 Traces & 20 Traces & 10 Traces & 10 Traces with 10 ms Undersampling   \\
         \hline
         Training Time & 3.74 s & 1.61 s & 0.899 s & 0.188 s \\
         Per-sample Prediction Time & 83.67$\mu$s & 68.8$\mu$s & 62.34$\mu$s & 54.4$\mu$s \\
         \hline
    \end{tabular}
    \label{table:4:Time_stats}
\end{table*}

\begin{figure}[!h]
\centering
\includegraphics[width=8cm]{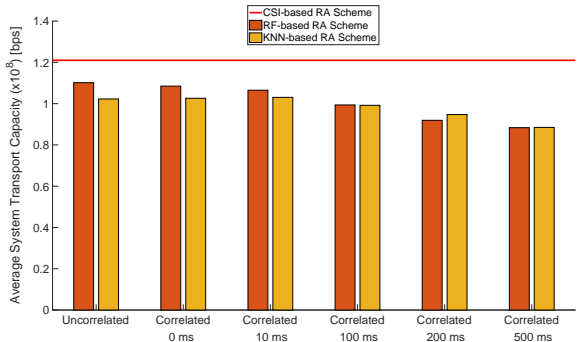}
\caption{System performance for 8$\times$2 MIMO system with uncorrelated dataset and correlated  dataset having 10 traces for $\rho~\leq~$0.05/m$^2$ with erroneous positions. The undersampling in the correlated channel dataset  varies from 0 to 500 ms.}
\label{fig:20:result_correlated_erroneous_pos_undersample}
\end{figure}

All of these results for correlated channel datasets are generated with the assumption that the terminal position is accurately known by the system. 
But in reality, the estimated position is inaccurate, involving some degree of error. 
We now assume that the acquired estimates of terminal position are erroneous, with the error modelled as a zero-mean Gaussian and a variance determined by $\sigma =$ 0.4 m. 
Fig.~\ref{fig:20:result_correlated_erroneous_pos_undersample} shows the resulting performance when erroneous positions are used for training the machine learning frameworks using 10 traces in the correlated dataset, with different rates of undersampling. 
The performance is quite robust, even when a small undersampling rate is applied, compared to the uncorrelated channels' dataset. 
The same performance behavior between RF and KNN is observed here as with the perfect positions' data: KNN performs at par with RF as the number of samples in the training dataset decreases. 
    
In addition to the performance evaluation, we also calculated the training time required by RF, as well as the prediction time per sample, when different number of traces and undersampling rate is considered for generating the real-time dataset. 
Table~\ref{table:4:Time_stats} presents the time required to train the RF model and the prediction time per-sample it takes, for different sizes of the dataset with correlated channel. 
It takes less than a second to train the RF model, with $(\Omega_{\textrm{t}}, \Omega_{\textrm{d}}) = (50, 12)$, for a dataset size of $\sim$10,000 samples that needs less than 1 kB of memory storage. The prediction time per-sample is also also very small, significantly lesser than the transmission time interval $T_\mathrm{f} =$ 0.2 ms of the system. All these results point towards the feasibility of implementation of the proposed coordinates-based resource allocation scheme using machine learning frameworks: The system can switch from the training-based mode to the position-based mode within a minute, and does not need to be re-trained frequently since the performance of the proposed scheme is quite stable compared to the CSI-based scheme as shown by the results presented in this work. 

\section{Conclusion}
\label{sec:6:conclusion}
This work presented a detailed design of coordinates-based resource allocation scheme using machine learning frameworks. 
We used supervised machine learning to learn the relationship between the terminal's position coordinates and the associated resource allocation to maximize the transport capacity of the system. 
A performance adjusted accuracy metric was introduced to determine the prediction performance of the learning frameworks with different dataset formulations, based on which the dataset formulation $\pmb{D}_2$ was found to be the best. 
The average system transport capacity is used for performance comparison between the proposed coordinates-based resource allocation scheme, a simple geometry-based scheme and a legacy CSI-based resource allocation scheme. 
The results show that the proposed scheme outperforms the geometry-based scheme by a significant margin, irrespective of the antenna configuration considered in the system. 
The proposed scheme performs consistently well in comparison to the CSI-based resource allocation scheme and is robust to the different stochastic variations in the system. 
These results are consistent when realistic system set up is considered, where the samples used for training the machine learning frameworks have correlated channel. 
Surprisingly, the proposed scheme needs a training dataset with samples collected over a couple of seconds to achieve a transport capacity of 95\% compared to the CSI-based resource allocation scheme. 
In terms of the system resources, the learnt model using random forest algorithm needs less than one second to train and requires less than 1 kB of memory storage for predicting an appropriate resource allocation for a given terminal's position estimate. 

The results from this study are very encouraging to establish the feasibility of coordinates-based resource allocation for the communication link between an individual base station and a mobile terminal. 
In future work, we will extend our investigation by applying the proposed coordinates-based resource allocation scheme to an interference-limited system, i.e. multiple base stations serving multiple mobile terminals. 
The interference posed by both the interfering terminals as well as by the neighbouring base stations will bring up new challenges for designing the coordinates-based resource allocation scheme. 
It will also be interesting to see how the training time as well as the size of the machine learning model scales with the multiple-transmitters, multiple-users system.


%



\section*{Acknowledgment}

A major part of the computations in this work was performed on resources provided by the Swedish National Infrastructure for Computing (SNIC) at PDC, KTH.

\vfill

\end{document}